\documentclass[10pt,twocolumn,letterpaper]{article}

\usepackage[pagenumbers]{cvpr}

\usepackage{placeins}

\usepackage{iftex}
\ifPDFTeX
  \usepackage[T1]{fontenc}
\else
  \usepackage{fontspec}
\fi

\usepackage[protrusion=true,expansion=false]{microtype}

\definecolor{linkblue}{rgb}{0.21,0.49,0.74}
\usepackage[breaklinks,colorlinks,allcolors=linkblue]{hyperref}

\title{SAGA: A Surgical Agentic Generation Architecture for Perception-in-the-Loop Controllable Video Generation}

\author{
  Qirui Zhong$^{1}$ \quad Juncheng Wang$^{2}$ \quad Shujun Wang$^{2}$ \quad Cheng Xue$^{1}$\textsuperscript{*}\\
  $^{1}$Southeast University \quad $^{2}$Hong Kong Polytechnic University\\
  {\tt\small qiruizhong@seu.edu.cn}
}
\date{}

\begin{document}
\maketitle

\begingroup
\renewcommand{\thefootnote}{*}
\footnotetext{Corresponding author: Cheng Xue.}
\endgroup

\begin{abstract}

Recent advances in video generation provide new opportunities for surgical video
synthesis, yet controllable surgical video generation remains challenging due to
the stochastic nature of generative models and the strict requirements on
instrument manipulation and surgical consistency. Existing approaches typically
follow an open-loop paradigm, assuming that generated videos faithfully execute
the given motion commands, which may lead to visually plausible but
uncontrollable surgical motions. In this work, we propose \textbf{SAGA}, a \textbf{S}urgical \textbf{A}gentic
\textbf{G}eneration \textbf{A}rchitecture for perception-in-the-loop controllable
video generation. SAGA decouples frozen video synthesis from perception-driven
control through an agent that coordinates state estimation, verification,
refinement, and cross-segment feedback. Goal gating separates motion
requirements from structural ranking, while observed outcomes guide subsequent
commands without generator training. We further introduce \textbf{SurgControl-Bench}, a benchmark that jointly
evaluates responsiveness to trajectory inputs, image-space geometric
consistency, and instrument appearance consistency.
Experiments across four frozen generator families characterize the effects and trade-offs of search, verification, refinement and feedback. Independent TAPIR evaluation shows lower endpoint error with best-of-six selection across all four hosts, while DINOv2 measurements and geometric diagnostics characterize verification trade-offs and refinement gains on the shared evaluation set.

\end{abstract}

\begin{figure}[!t]
    \centering
    \includegraphics[width=\columnwidth]{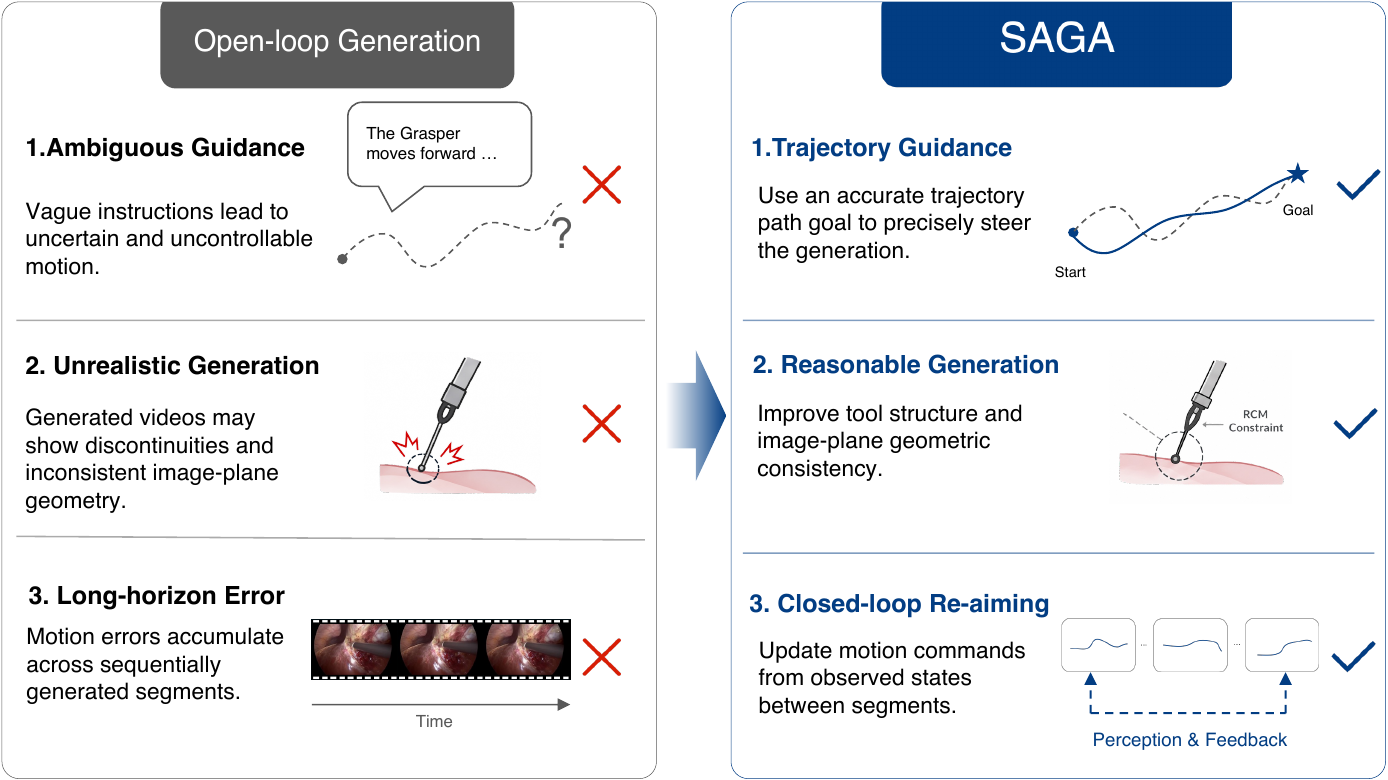}
    \caption{\textbf{From open-loop generation to SAGA.} SAGA combines trajectory
    guidance, checks of tool structure and image-plane geometry, and closed-loop
    re-aiming to improve command following across segments.}
    \label{fig:teaser}
\end{figure}

\section{Introduction}
\label{sec:intro}

Recent advances in video generation have enabled the synthesis of increasingly
realistic and diverse visual content. However, achieving reliable controllability
remains a fundamental challenge, as generated videos may deviate from the
specified conditions due to the stochastic nature of generative models. This
challenge becomes particularly critical in surgical scenarios, where generated
videos are expected not only to appear realistic but also to accurately follow
instrument manipulation and preserve surgical constraints. Unreliable synthesis
may lead to unrealistic procedure simulation and limit the utility of surgical
video generation for training, planning, and education.
These limitations take three linked forms: ambiguous guidance yields uncertain
motion; visually plausible generation may violate physical laws, RCM constraints,
or tool integrity; and open-loop errors accumulate over long horizons, causing
drift and quality degradation.

A key limitation of existing controllable video generation lies in its open-loop
assumption: the provided motion condition is treated as a guaranteed execution
command. However, generated motion does not necessarily correspond to the
intended surgical action. A video may achieve the target instrument location
while violating fundamental surgical properties, such as trocar
remote-center-of-motion (RCM) constraints, instrument continuity, or realistic
tool--tissue interactions. Therefore, controllability cannot be evaluated solely
by target reaching accuracy. Existing benchmarks assess visual fidelity~\cite{huang2025vbench++}, general physical plausibility~\cite{bansal2025videophy,motamed2025generative}, and trajectory accuracy alongside video quality~\cite{li2025magicmotion}. Surgical evaluation additionally requires instrument-level command-response tests and checks of tool structure and image-space geometric consistency. Figure~\ref{fig:teaser} illustrates these three limitations and SAGA's corresponding mechanisms: explicit trajectory guidance, image-based checks
of tool integrity and geometric consistency, and closed-loop feedback
correction of accumulated motion errors.

To address this challenge, we introduce \textbf{SAGA}, a
\textbf{S}urgical \textbf{A}gentic \textbf{G}eneration \textbf{A}rchitecture for
perception-in-the-loop controllable video generation. Instead of treating video
generation as a one-shot synthesis process, SAGA formulates it as a
closed-loop perception and decision-making process. Given a frozen
trajectory-conditioned generator, SAGA observes generated outcomes, evaluates
their surgical reliability, and adaptively refines subsequent generation through
feedback. Without additional training, SAGA integrates surgical state
perception, multi-dimensional verification, and perception-guided refinement to
transform frozen generators into controllable surgical video generation systems.

Furthermore, we introduce \textbf{SurgControl-Bench}, a benchmark for evaluating
controllable surgical video generation beyond conventional visual quality
metrics. SurgControl-Bench tests command responsiveness with paired trajectory
inputs, alongside measurements of instrument appearance consistency and
image-space geometric consistency. Here, RCM is assessed through image-space axis--pivot consistency, while the remaining geometry and tool-integrity measures provide complementary structural diagnostics. Joint motion and camera diagnostics help distinguish instrument motion responsive to the command from apparent motion associated with camera changes.

Our contributions are summarized as follows:

\begin{itemize}
    \item We propose \textbf{SAGA}, an agentic framework that decouples frozen
    video synthesis from perception-driven control. Modular state estimation,
    goal-gated structural verification, refinement, and cross-segment feedback
    improve controllability and consistency without generator training.

    \item We introduce \textbf{SurgControl-Bench}, a benchmark dedicated to
    evaluating controllable surgical video generation, measuring command following,
    image-space geometric consistency, and instrument appearance.

    \item We evaluate four frozen generator families with decision and evaluation
    signals decoupled: CoTracker2 drives the agent; TAPIR and DINOv2 independently
    assess fixed outputs and characterize component gains and trade-offs.

\end{itemize}

\begin{figure*}[t]
    \centering
    \includegraphics[width=0.98\textwidth]{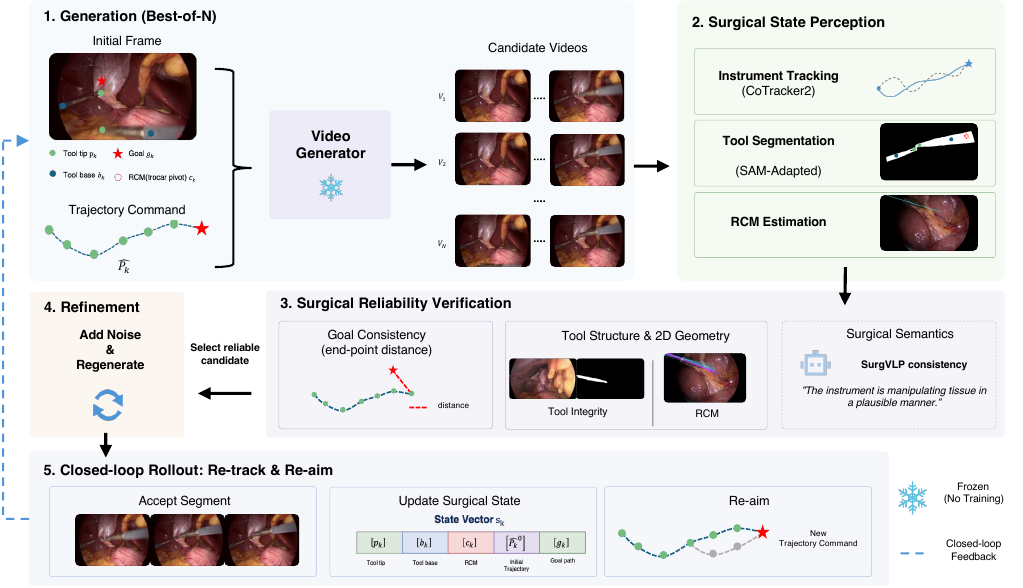}
    \caption{\textbf{SAGA overview.} A frozen generator samples trajectory-conditioned
    candidates. CoTracker2 and tool masks guide verification and refinement using
    motion and image-plane geometry. The retained segment updates the state for
    re-tracking and re-aiming. The dashed semantic branch is optional and disabled
    in the main-table experiments.}
    \label{fig:overview}
\end{figure*}

\section{Related Work}

\subsection{Surgical Video Generation}

Dedicated surgical generators~\cite{chen2025surgsora,koju2025surgical,li2024endora}
support education and analysis, while general-purpose world
models~\cite{he2025surgworld,lin2024world,nvidia2025cosmos,yu2024orbit} have been
adapted for surgical robot training and action generation. Their evaluations
emphasize visual fidelity, leaving command following and axis--pivot consistency
largely unexamined.

\subsection{Controllable Video Generation}

Recent work improves controllability through structured representations,
reinforcement learning, and subject decoupling~\cite{chu2026wan,xu2026smrabooth,xu2026disco}.
Our frozen hosts cover latent-trajectory guidance~\cite{wan2025wan,chu2026wan},
trajectory-token diffusion~\cite{zhang2025tora,yang2025cogvideox},
drag-based control~\cite{wu2024draganything,blattmann2023stable}, and
flow-based control~\cite{li2025image,guo2024animatediff}.
PoseTraj~\cite{ji2025posetraj} learns pose-aware trajectory control through
3D box supervision and camera-disentangled fine-tuning.
MagicMotion~\cite{li2025magicmotion} progressively learns control with
masks, boxes, and sparsely annotated boxes, and introduces MagicBench
to evaluate trajectory accuracy and video quality across object counts.
SAGA guides frozen generators using observations of instrument motion
and surgical structural checks.

\subsection{Agentic Frameworks}

Agentic systems use tools and feedback for iterative decision making,
including medical diagnosis and planning.
Test-time scaling allocates additional inference computation to
candidate search~\cite{brown2024large,guo2025deepseek,muennighoff2025s1,ma2025inference,he2025scaling}.
Video-T1~\cite{liu2025videot1} studies verifier-guided candidate selection
and Tree-of-Frames search for video generation.
SAGA uses observed instrument states to update motion commands between
segments and couples this feedback with checks of tool structure and
image-plane axis--pivot consistency for frozen trajectory-conditioned
generators.

\section{Methods}

\subsection{Overview}

Frozen generators can realize different motions under the same trajectory and
accumulate deviations over long horizons. \textbf{SAGA} (\textbf{S}urgical
\textbf{A}gentic \textbf{G}eneration \textbf{A}rchitecture) makes generation
perception-in-the-loop: given $I_0$ and $g$, it perceives the realized state,
re-aims future commands, verifies candidates, and refines unreliable executions,
without updating the host generator. Figure~\ref{fig:overview} summarizes this
generation--perception loop.

Decoupling assigns visual synthesis to the frozen host and control decisions
to the agent, which uses observed instrument states to select, revise, and
continue generation. Refinement revises a retained segment; re-aiming updates
subsequent commands. This decision structure is shared across host interfaces.

\subsection{Surgical State Perception and Closed-loop Planning}

SAGA observes the instrument motion realized by a frozen generator and
updates subsequent commands from this perceived state, closing the gap
between trajectory conditioning and execution.

Given an initial surgical context $I_0$ and a target motion goal $g$, SAGA
generates surgical videos in sequential segments. At generation step $k$, the
surgical perception module estimates the current execution state from the
generated visual context:

\begin{equation}
s_k =
[p_k,b_k,c_k,\bar{P}_k,g_k],
\end{equation}

where $p_k$ denotes the tracked instrument reference position, $b_k$ represents
the instrument configuration, $c_k$ describes the trocar pivot constraint,
$\bar{P}_k$ denotes the observed instrument trajectory, and $g_k$ represents
the remaining motion goal.

For a rollout of $S$ segments, closed-loop re-aiming distributes the remaining
displacement from the observed position $p_k$ to the final target $g$ over the
remaining segments. The open-loop baseline uses a fixed segment displacement:
\begin{equation}
\begin{aligned}
u_k^{\mathrm{closed}}&=\frac{g-p_k}{S-k},\\
u_k^{\mathrm{open}}&=\frac{g-p_0}{S},
\end{aligned}
\label{eq:reaim}
\end{equation}
where $k=0,\ldots,S-1$ and $u_k$ is the total displacement command for one
segment, with unit feedback gain. After each segment, SAGA re-tracks the retained
output to update $p_{k+1}$ and recomputes the next closed-loop command. For WAN,
$p_k$ is the mean of the tool-query positions; per-query trajectories are
linearly interpolated and clipped to the image bounds. The frozen generator
then samples candidates using the resulting command $u_k$:

\begin{equation}
v_k^{(i)}\sim\mathcal{W}(I_k,u_k,\xi_i),
\end{equation}

where $\xi_i$ denotes the sampling noise and $v_k^{(i)}$ represents the generated
candidate segment.

Unlike open-loop generation that relies solely on the predefined motion
condition, SAGA continuously incorporates generated outcomes as observations and
updates subsequent actions based on the executed surgical state. This agentic
perception loop enables reliable generation through subsequent candidate
verification and feedback refinement.

\subsection{Surgical Reliability Verification}

SAGA verifies sampled candidates using goal consistency, tool structure,
and image-space geometry.

Given the updated command $u_k$, the frozen generator $\mathcal{W}$ produces
multiple candidate segments:

\begin{equation}
\mathcal{V}_k=\{v_k^{(i)}\}_{i=1}^{N},
\end{equation}

where each candidate is generated with different sampling noise. SAGA
re-perceives each candidate and evaluates goal consistency, tool structure,
and image-space geometry. Semantic compatibility is additionally assessed
when semantic verification is enabled.

Goal consistency is measured using the inference-time tracker. A relative goal gate retains candidates within a specified tolerance of the lowest candidate error, after which the enabled verification signals rank the retained candidates. This gate is distinct from the absolute error threshold used to report endpoint success.


The reliability score combines geometric proxies with an optional semantic term:

\begin{equation}
Q_i=
Q_{\mathrm{geo}}^{(i)}
+\lambda_{\phi}\phi_i ,
\end{equation}

where $Q_{\mathrm{geo}}^{(i)}$ represents the configured geometric consistency score and $\phi_i$ is an optional semantic compatibility score. The active terms and weights are specified for each experiment. The candidate with the highest score among goal-consistent candidates is selected:

\begin{equation}
i^\star=\arg\max_{i\in\mathcal{G}_k}Q_i,
\qquad \hat v_k=v_k^{(i^\star)}.
\end{equation}
Here $\mathcal G_k$ contains the indices of candidates that pass the relative
goal gate.

For the fixed candidate comparison, Bo6 selects the candidate with the lowest inference-time goal error. Ver. retains candidates within 8 pixels of that minimum and minimizes the RCM residual plus 0.1 times the fragmentation measure. This configuration does not use semantic scores. Independent TAPIR and DINOv2 measurements are computed only after the policy outputs are fixed.

Goal gating constrains target deviation, while verification ranks candidates using the configured proxy criteria. Improvements in measurements excluded from these criteria are assessed separately on fixed policy outputs.

\subsection{Perception-guided Surgical Refinement}

Proxy-based candidate selection can leave residual motion errors.
SAGA therefore introduces perception-guided refinement to adjust the selected
generation using feedback from the estimated instrument state.

Starting from $v^{(0)}=\hat v_k$, SAGA uses partial-renoise editing and restart
sampling~\cite{meng2022sdedit,xu2023restart} to generate proposals from the
currently retained segment $v^{(r)}$:
\begin{equation}
v'\sim\mathcal{W}\bigl(v^{(r)},\tilde u^{(r)},\rho\bigr),
\end{equation}
where $\tilde u^{(r)}$ is the proposal motion command. All four generators use
partial-renoising strength $\rho=0.7$. Each stage evaluates at most four
proposals within the matched candidate budget.

Let $D(v)$ be the endpoint goal distance measured by the inference-time tracker,
and $F(v)$ the mean per-frame number of foreground connected components in the
tool masks. A proposal replaces the retained segment if it reduces the current
goal distance by more than $10^{-3}$ pixels, or if it reduces fragmentation
while keeping its goal distance at most $\delta$ pixels above the initial
distance $D(v^{(0)})$. We use $\delta=8$ pixels in the main-table experiments,
with the initial-distance reference fixed throughout refinement. The final
output is $v_k=v^{(R)}$ after $R\le4$ proposals. The complete acceptance rule
is given in the supplementary material. TAPIR and DINOv2 measurements are
computed after policy outputs are fixed.

After refinement, the accepted segment is incorporated into the surgical
context:

\begin{equation}
I_{k+1}=\mathrm{UpdateContext}(I_k,v_k),
\end{equation}

and the surgical state is updated for the next generation step:

\begin{equation}
s_{k+1}=\mathrm{Perceive}(I_{k+1},g).
\end{equation}

This refinement loop uses feedback from the estimated instrument state
to reduce motion deviations and limit drift during multi-segment generation.

\subsection{Closed-loop Surgical Video Rollout}

For long procedures, SAGA updates the context and re-estimates the state after
each accepted segment, then re-aims the remaining command. Concatenating these
feedback-guided segments yields long-horizon generation without updating the
frozen generator.

\begin{table*}[t]
\centering
\caption{\textbf{SurgControl-Bench cross-host results.} WAN, DA, and IC denote WAN2.2,
    DragAnything, and ImageConductor; Bo6/Ver./Ref. denote best-of-six selection, verification,
    and refinement/retry. Ver./Ref. use matched candidate counts.
    Goal is TAPIR endpoint error (px); S10/S20 are visibility-aware
    success rates at 10/20 px (\%). $\Delta$DINO is the mean paired change in first-frame
    tool-ROI DINOv2 similarity relative to Single. Bold/underline mark the best/second-best
    values within each host; a dash marks the Single reference.}
\label{tab:benchmark}
\normalsize
\setlength{\tabcolsep}{3.5pt}
\begin{tabular*}{\textwidth}{@{\extracolsep{\fill}}llrrrrrrrrrr@{}}
\toprule
& & \multicolumn{3}{c}{Control}
    & \multicolumn{4}{c}{Geometry}
    & \multicolumn{2}{c}{Appearance}
    & Consistency \\
\cmidrule(lr){3-5}
\cmidrule(lr){6-9}
\cmidrule(lr){10-11}
\cmidrule(lr){12-12}
Host & Policy 
    & Goal$\downarrow$ & S10$\uparrow$    & S20$\uparrow$     
    & Cam.$\downarrow$ & FE$_{\mathrm{2D}}\downarrow$ & FE$_{\mathrm{true}}\downarrow$ & Depth$\downarrow$  
    & Frag.$\downarrow$ & $\Delta$DINO$\uparrow$  
    & RCM$\downarrow$  \\  
\midrule
%
WAN  & Single 
    & 58.4             & 2.2              & 9.4               
    & 0.69 & 33.5             & 31.7             & .112              
    & 3.6              & --         
    & .032             \\  
     & $+$Bo6 
    & \underline{28.8} & \underline{11.9} & \underline{36.3}  
    & \underline{0.67}           & 21.4    & 28.3    & \underline{.111}  
    & 3.6              & $+0.0036$  
    & .033             \\  
     & $+$Ver.
    & 30.2             & 10.6             & 32.5              
    & \underline{0.67}           & \underline{18.3} & \underline{25.7} & \underline{.111}  
    & \underline{3.0}  & \underline{$+0.0039$}  
    & \underline{.030} \\  
     & $+$Ref.
    & \textbf{22.2}    & \textbf{25.0}    & \textbf{52.5}     
    & \textbf{0.59}    & \textbf{18.0}             & \textbf{25.6}             & \textbf{.105}     
    & \textbf{2.4}     & $\mathbf{+0.0045}$         
    & \textbf{.028}    \\  
\midrule
%
Tora & Single 
    & 32.3             & 13.9             & 41.5              
    & 0.51 & 22.9             & 23.6             & .112     
    & 5.1              & --         
    & .020             \\  
     & $+$Bo6 
    & \underline{13.6} & \underline{54.4} & \underline{80.6}  
    & 0.50             & 17.4     & 21.6    & .111              
    & 3.2  & $+0.0034$  
    & .020             \\  
     & $+$Ver.
    & 15.1             & 40.0             & 78.8              
    & \underline{0.47} & \underline{9.6}  & \underline{16.6} & \underline{.109}  
    & \underline{3.0}     & \underline{$+0.0176$}  
    & \underline{.019} \\  
     & $+$Ref.
    & \textbf{7.5}     & \textbf{66.9}    & \textbf{100.0}    
    & \textbf{0.38}    & \textbf{9.2}             & \textbf{16.4}             & \textbf{.107}              
    & \textbf{2.5}              & $\mathbf{+0.0213}$          
    & \textbf{.013}    \\  
\midrule
%
DA   & Single 
    & 15.5             & 38.2             & 75.9              
    & 0.96             & 16.3              & 20.0 & .136              
    & 3.9  & --         
    & .016             \\  
     & $+$Bo6 
    & \underline{6.4}  & \underline{83.1} & \underline{98.1}  
    & 0.97             & 6.7     & 19.8             & .130  
    & \underline{3.3}              & $+0.0035$   
    & .016             \\  
     & $+$Ver.
    & 9.6              & 58.8             & 95.6              
    & \underline{0.94} & \underline{4.8}  & \underline{19.2}    & \underline{.127}    
    & \underline{3.3}     & \underline{$+0.0081$}  
    & \underline{.012} \\  
     & $+$Ref.
    & \textbf{5.1}     & \textbf{95.0}    & \textbf{100.0}    
    & \textbf{0.75}    & \textbf{3.9}              & \textbf{19.1} & \textbf{.120}             
    & \textbf{2.3}  & $\mathbf{+0.0121}$         
    & \textbf{.011}    \\  
\midrule
%
IC   & Single 
    & 22.9             & 17.6             & 52.6              
    & 1.06 & 11.4              & 19.4 & .335  
    & 28.5             & --         
    & .051             \\  
     & $+$Bo6 
    & \underline{9.3}  & \underline{60.0} & \underline{96.3}  
    & 1.04             & 6.1    & 18.5             & \underline{.255}     
    & 26.6 & $+0.0031$  
    & .049             \\  
     & $+$Ver.
    & 11.3             & 41.3             & 92.5              
    & \underline{1.00} & \underline{3.5}  & \underline{18.3}             & \textbf{.246}     
    & \underline{26.3}    & $\underline{+0.0064}$  
    & \underline{.046} \\  
     & $+$Ref.
    & \textbf{4.0}     & \textbf{95.0}    & \textbf{100.0}    
    & \textbf{0.71}    & \textbf{3.2}              & \textbf{14.8}    & \textbf{.246}              
    & \textbf{15.0}             & $\mathbf{+0.0096}$         
    & \textbf{.038}    \\  
\bottomrule
\end{tabular*}
\end{table*}

\subsection{Implementation Details}

SAGA requires no additional training or fine-tuning of the video generator.
The generator and all auxiliary models remain frozen during generation.
Auxiliary perception models are pretrained or adapted offline, as described
below. Planning, verification, and refinement are performed at inference time
through the generator's conditioning interface.

CoTracker2~\cite{karaev2024cotracker} provides the agent's tracking feedback for selection, proposal acceptance, and trajectory correction. TAPIR evaluates fixed outputs in the single-segment, rollout, and candidate-budget experiments, independently of all agent decisions.
We use a SAM-adapted surgical tool segmenter~\cite{kirillov2023segment,ma2024segment,wu2025medical} to obtain tool masks for identity consistency, fragmentation, and axis--pivot constraints.

The segmenter is adapted using CholecSeg8k masks~\cite{hong2020cholecseg8k};
a YOLO26-based motion slicer, trained on CholecTrack20 tracks~\cite{nwoye2025cholectrack20,jocher2026ultralytics},
selects instrument-active source windows. Camera-motion and depth diagnostics use RAFT~\cite{teed2020raft}
and Depth Anything 3~\cite{lin2025depth}, respectively. 
For experiments that enable semantic verification, we use SurgVLP~\cite{yuan2025learningmultimodalrepresentationswatching} to score compatibility with surgical descriptions. The inference-time perception and verification modules remain frozen during generation. TAPIR and DINOv2 are used only for post-hoc evaluation of the fixed generated outputs.

Cam. is the camera-to-observed-motion magnitude ratio, with accumulated
observed tool-path length as its denominator; it can exceed one.
RCM is the RMS image-plane residual of PCA tool-shaft axes relative to a
least-squares shared pivot, normalized by the scoring-mask image width $W$.
The widths are 832, 720, 576, and 384 pixels for WAN, Tora, DragAnything, and
ImageConductor, respectively. Lower residuals indicate greater axis--pivot
consistency. Full metric formulas, host-specific proposal commands, and
axis-fitting details are provided in the supplementary material.
$\mathrm{FE}_{\mathrm{2D}}$ and $\mathrm{FE}_{\mathrm{true}}$ denote raw and
camera-compensated instrument-following errors, respectively.
 Depth measures
the mean absolute change between spatially standardized consecutive depth
maps, and Frag. reports $F(v)$.

During inference, SAGA iteratively performs perception, verification, and
perception-guided refinement, enabling controllable surgical video generation without additional training.

\section{SurgControl-Bench}

Existing evaluations of surgical video generation mainly focus on visual
fidelity, which cannot determine whether generated videos genuinely follow
specified surgical actions. A generator may produce realistic-looking surgical
scenes while ignoring the input motion command or achieving apparent control
through camera movement rather than instrument manipulation.

\begin{figure*}[t]
    \centering
    \includegraphics[width=0.98\textwidth]{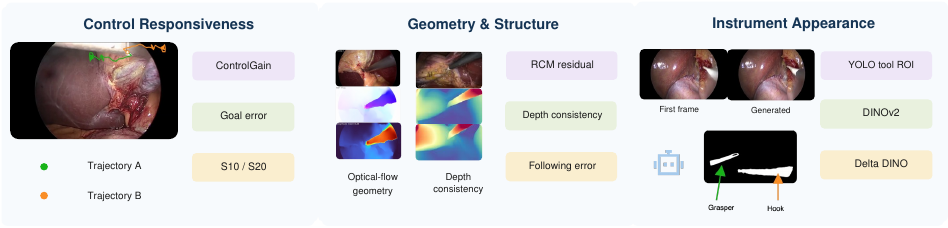}
    \caption{\textbf{SurgControl-Bench.} Fixed policy outputs are assessed for control
    responsiveness, geometric and structural consistency, and instrument appearance.
    TAPIR measures endpoint error and visibility-aware success at 10/20 px;
    DINOv2 measures first-frame appearance similarity on YOLO-detected tool regions,
    with $\Delta$DINO reported relative to Single. TAPIR and DINOv2 scores are excluded
    from selection and refinement.}

    \label{fig:benchmark}
\end{figure*}

To address this limitation, we introduce \textbf{SurgControl-Bench}, a benchmark
for evaluating controllable surgical video generation beyond visual quality.
SurgControl-Bench evaluates generated videos from three complementary
perspectives: (1) \textbf{control responsiveness}, measuring whether instrument
motion follows the specified command; (2) \textbf{surgical consistency},
assessing geometric structure, instrument integrity, and image-space
axis--pivot consistency; and (3) \textbf{instrument appearance}, measuring
visual consistency of tool regions.
Figure~\ref{fig:benchmark} summarizes these three evaluation axes.

\subsection{Evaluation Protocol}

\textbf{Independent Motion Evaluation.}
We evaluate the fixed generated outputs with TAPIR~\cite{doersch2023tapir}, independently of the CoTracker2 feedback used within the agent. Within each experiment, compared policies share the query definition, coordinate mapping, visibility criteria, and aggregation rules. For the main-table comparison, we track the same 11--64 source-frame instrument points across policies and map their coordinates to an $832\times480$ reference frame. Goal is the raw distance between the mean tracked endpoint and the target, averaged over all clips. A query endpoint is visible when its TAPIR confidence exceeds 0.5. S10 and S20 require both a goal distance within 10 and 20 pixels, respectively, and visibility of at least half the query endpoints; all 160 clips remain in the success-rate denominator for each host.

\textbf{Instrument Appearance Evaluation.}
We assess instrument appearance using DINOv2~\cite{oquab2024dinov2} features extracted from YOLO-detected tool regions. We measure cosine similarity to the first-frame tool region. For Bo6, Ver., and Ref., $\Delta$DINO is the mean paired difference from Single over all 160 clips in the shared evaluation set. Positive values indicate higher appearance similarity than Single; the Single reference is shown as a dash. DINOv2 scores are computed only after the policy outputs are fixed.

\textbf{Evaluation with Paired Trajectories.}
To test whether trajectory conditioning affects generated motion, we vary the
input trajectory while holding the other generation conditions fixed. Here,
counterfactual denotes an alternative trajectory input in a controlled
generation test; the measured capability is responsiveness to that input.

For each initial surgical context, we generate two videos with opposite motion
commands while keeping the initial condition, sampling noise, and candidate
budget identical.
Given a command trajectory $A$, its counterfactual counterpart is constructed
as:

\begin{equation}
B_t=2A_0-A_t .
\end{equation}

With the initial context and sampling noise fixed, we reverse the command
trajectory to measure how generated instrument motion responds to the control
input. We define \textbf{ControlGain} as the ratio of the separation between the
generated trajectories to that between the commanded trajectories:

\begin{equation}
\mathrm{ControlGain}
=
\frac{
\frac{1}{T}\sum_t
\|\hat p_t^A-\hat p_t^B\|_2
}
{
\frac{1}{T}\sum_t
\|A_t-B_t\|_2+\epsilon
}.
\end{equation}

Here $T$ is the number of frames, $\hat p_t^A$ and $\hat p_t^B$ are tracked
instrument reference positions in the two generated videos, and $\epsilon$
is a small stabilizer.
ControlGain measures the magnitude of the response to command reversal.
Values near zero indicate weak trajectory separation, values near one indicate
separation comparable to that of the commands, and values above one indicate
amplified separation. We interpret it jointly with following errors and endpoint
success to assess both command responsiveness and trajectory alignment.

\textbf{Multi-axis Evaluation.}
Beyond command responsiveness, SurgControl-Bench evaluates tool structure,
image-space geometric consistency, and instrument appearance. We report
camera-compensated following errors, depth consistency, tool-mask fragmentation,
axis--pivot residuals, and tool-ROI DINOv2 similarity. Together with the control
metrics, these measurements characterize motion, structure, and appearance
in generated surgical videos.

\section{Experiments}

\subsection{Experimental Setup}

We evaluate SAGA as a training-free inference-time framework on four frozen
surgical video generators: WAN2.2-TI2V-5B~\cite{wan2025wan} with
WAN-Move~\cite{chu2026wan}, Tora~\cite{zhang2025tora},
DragAnything~\cite{wu2024draganything}, and ImageConductor~\cite{li2025image}.
These generators represent different controllable video synthesis paradigms,
including trajectory-based motion guidance and interactive visual control.
All video generators remain frozen during evaluation. SAGA accesses their
conditioning interfaces without generator fine-tuning; auxiliary perception
and verification models also remain frozen during generation.

All four main-table policies---Single, Bo6, Ver., and Ref.---use the same 160 Cholec80~\cite{endonet} clips from 80 source videos per host, with identical initial frames, motion commands, and evaluation metrics. Single averages six seeds; Bo6 selects by goal error, and Ver. adds structural verification. Ver. and Ref. generate the same total number of candidates per clip, including all refinement/retry proposals. TAPIR and DINOv2 evaluate fixed policy outputs. Multi-segment rollouts are reported separately.

Table~\ref{tab:benchmark} reports TAPIR Goal, visibility-aware S10/S20, and tool-ROI $\Delta$DINO relative to Single. Camera motion, following error, depth, fragmentation, and RCM residual are reported as perception-based diagnostics. Main-table comparisons describe the reported summary values; statistical tests for separate experiments are identified with their corresponding results.

\subsection{Independent Motion and Appearance Evaluation}

Table~\ref{tab:benchmark} evaluates all four policies on the shared 160-clip set with TAPIR and DINOv2, whose scores are excluded from policy selection. Bo6 lowers mean TAPIR Goal from 58.4 to 28.8 pixels on WAN, 32.3 to 13.6 on Tora, 15.5 to 6.4 on DragAnything, and 22.9 to 9.3 on ImageConductor. Visibility-aware S20 increases from 9.4\% to 36.3\%, 41.5\% to 80.6\%, 75.9\% to 98.1\%, and 52.6\% to 96.3\%, respectively.

Verification changes the endpoint--appearance trade-off. Relative to Bo6, Ver. increases Goal by 1.4, 1.5, 3.2, and 2.0 pixels on WAN, Tora, DragAnything, and ImageConductor, respectively, with lower S10/S20 on all four hosts. Its reported appearance gains relative to Single are larger: for example, $\Delta$DINO rises from +0.0034 to +0.0176 on Tora and from +0.0035 to +0.0081 on DragAnything. RCM residual decreases on all four hosts, while fragmentation decreases or stays unchanged. These summary results characterize a trade-off rather than uniform improvement in endpoint accuracy.

\begin{figure}[!t]
\centering
\includegraphics[width=\columnwidth]{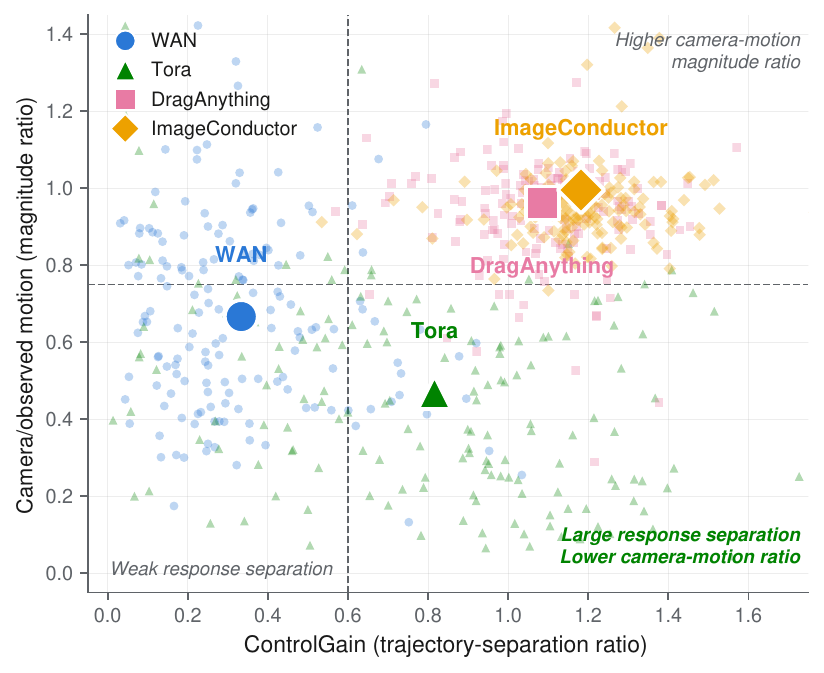}
\caption{\textbf{Motion diagnostics of generated videos.} ControlGain from TAPIR tracks versus the camera-to-observed-motion magnitude ratio. Paired trajectories test responsiveness to the input under matched initial context, sampling noise, and candidate budget. Small points denote clips and large points denote host means.}

\label{fig:controlgain}
\end{figure}

Figure~\ref{fig:controlgain} diagnoses generated videos via response separation and camera-motion magnitude. Paired inputs differ only in trajectory; initial context, noise, and candidate budget are matched. ControlGain measures responsiveness; following errors and endpoint success assess alignment.

\subsection{Closed-loop Re-aim for Long-horizon Generation}

Single-segment optimization cannot prevent error accumulation in long surgical
procedures. We compare open- and closed-loop rollouts using the same generator,
initial context, target command, four segments, and candidate budget on the
same 160 clips per host. CoTracker2 supplies state estimates for closed-loop re-aiming; TAPIR
independently evaluates both arms after their generated outputs are fixed,
using the same motion-evaluation protocol.

\begin{figure}[!t]
\centering
\includegraphics[width=\columnwidth]{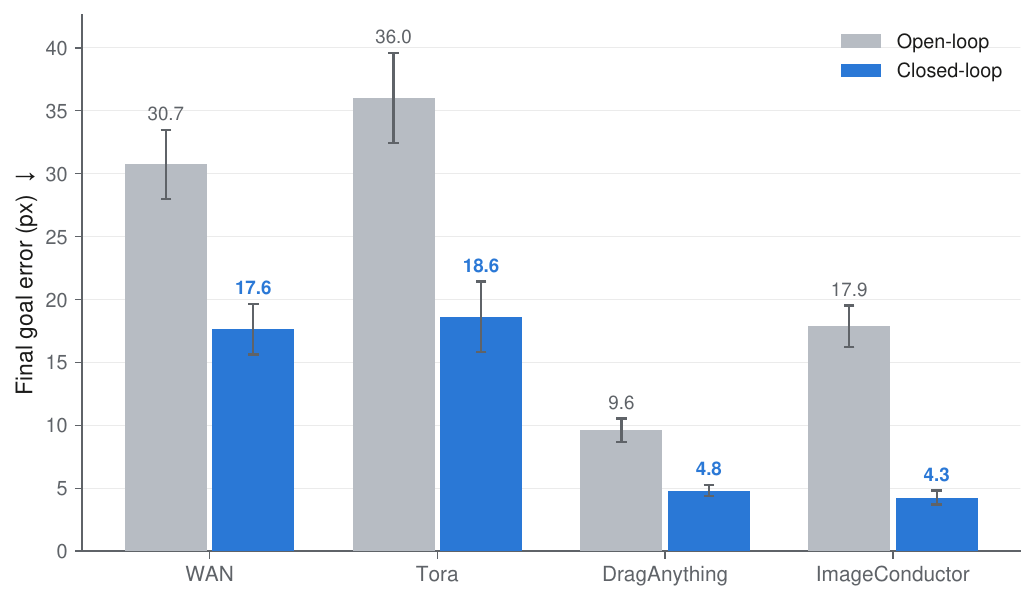}
\caption{\textbf{Closed-loop re-aim vs.\ matched open-loop rollout} (mean final
TAPIR goal error, px). CoTracker2 supplies agent feedback; TAPIR evaluates the fixed outputs.
Error bars indicate the standard error of the mean (SEM).
Both arms use four segments with matched sampling budgets within each host.
Re-aim lowers error by $43$--$76\%$ (paired Wilcoxon, all $p<10^{-4}$).}
\label{fig:closed_open}
\end{figure}

As shown in Figure~\ref{fig:closed_open}, closed-loop re-aim reduces final target
error by 43\%, 48\%, 50\%, and 76\% for WAN, Tora, DragAnything, and
ImageConductor, respectively (all $p<10^{-4}$). Both arms use matched candidate
budgets and identical TAPIR evaluation settings, assessing the effect of state
feedback under the matched rollout protocol.

\begin{samepage}
These results show that state feedback reduces final target error
in the evaluated multi-segment rollouts.\par
\end{samepage}

\subsection{Effect of Surgical Verification and Refinement}

Table~\ref{tab:benchmark} shows the endpoint--appearance trade-off: Ver. has
higher TAPIR Goal and lower S10/S20 than Bo6, but larger $\Delta$DINO
relative to Single on every host. On DragAnything, Goal rises from 6.4
to 9.6 pixels and S10 falls from 83.1\% to 58.8\%.

Table~\ref{tab:frontier} provides a separate goal-gate analysis.
The fixed-gate rows minimize RCM within the admitted candidates, whereas
\emph{policyG12} uses a standardized multi-criterion cost over goal error,
RCM, fragmentation, and mask-based identity consistency.
Its relative \emph{best+$12$} gate is distinct from the best+$8$ configuration
in the main candidate comparison and from an absolute success threshold.
This configuration lowers the RCM residual by 32.6\% while increasing goal
error by 13.3\%, illustrating the endpoint--geometry trade-off.

At matched candidate counts on the same 160 clips, Ref. lowers Goal
relative to Ver. from 30.2 to 22.2 pixels on WAN, 15.1 to 7.5 on Tora,
9.6 to 5.1 on DragAnything, and 11.3 to 4.0 on ImageConductor.
Its $\Delta$DINO relative to Single is +0.0045, +0.0213, +0.0121,
and +0.0096, respectively, all above Ver. Search, goal-gated verification,
and refinement together lower mean TAPIR Goal from Single by 62.0\%,
76.8\%, 67.1\%, and 82.5\%, respectively, with improved tool appearance
and image-plane axis--pivot consistency.

Refinement at matched candidate counts and independently evaluated rollout
and appearance results support within- and cross-segment decisions.
Reliability here comprises command following, tool appearance,
and image-plane geometric consistency.

\begin{table}[t]
\centering
\caption{\textbf{Separate verification goal-gate analysis.} ``Gate $x$ px'' uses a
relative \emph{best+$x$} endpoint tolerance; changes are relative to goal-only
selection. This 160-clip configuration study is separate from the main-table comparison.}
\label{tab:frontier}
\scriptsize
\setlength{\tabcolsep}{1.5pt}
\begin{tabular*}{\columnwidth}{@{\extracolsep{\fill}}lrrrr@{}}
\toprule
Policy & RCM $\Delta$ (\%) & $p_{\mathrm{RCM}}$ & Goal $\Delta$ (\%) & $p_{\mathrm{Goal}}$ \\
\midrule
Gate $6$~px          & $-34.5\%$ & $2{\times}10^{-4}$ & $+14.8\%$ & $0.30$ \\
Gate $12$~px         & $-42.5\%$ & ${<}10^{-4}$ & $+31.0\%$ & $0.006$ \\
Gate $18$~px         & $-46.5\%$ & ${<}10^{-4}$ & $+53.3\%$ & $5{\times}10^{-4}$ \\
\emph{policyG12} & $-32.6\%$ & $1.335{\times}10^{-5}$ & $\mathbf{+13.3\%}$ & $0.0457$ \\
\bottomrule
\end{tabular*}
\end{table}

\subsection{Test-time Scaling and Efficiency}

To assess budget--control scaling, each host uses 160 clips at the same
tested candidate counts $N$. CoTracker2 guides selection;
TAPIR evaluates the fixed outputs.

\begin{figure}[!t]
\centering
\includegraphics[width=\columnwidth]{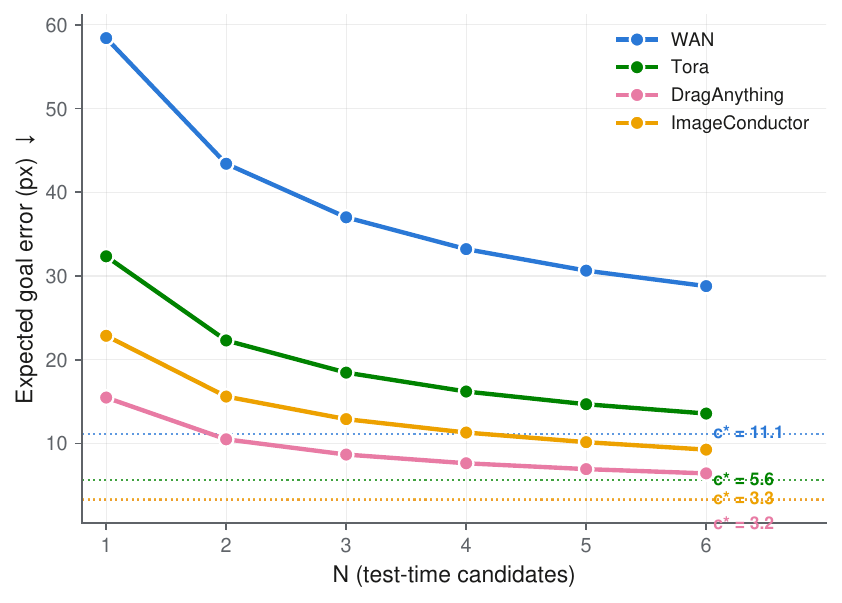}
\caption{\textbf{Empirical scaling over the tested candidate budgets.}
Expected TAPIR goal error versus $N$ for each frozen host. Dotted lines mark
the fitted offset $c^\star$; leave-one-budget-out prediction error is
$\leq0.23$~px.}
\label{fig:scaling}
\end{figure}

\begin{samepage}
As shown in Figure~\ref{fig:scaling}, increasing the number of candidates
consistently reduces target error but gradually reaches a saturation point.
\noindent This behavior can be modeled as:

\begin{equation}
\mathbb{E}[\mathrm{GoalError}|N]
=
c^\star+A N^{-\beta},
\end{equation}

where $N=1,\ldots,6$ is the tested candidate count, and $c^\star$, $A$,
and $\beta$ are the fitted offset, scale, and decay exponent within this range.
\end{samepage}

Within the tested range, leave-one-budget-out prediction error is at most
0.23 px. SAGA allocates more computation to difficult segments and stops
early on solved ones, achieving comparable controllability with up to
$3.3\times$ fewer candidate generations.

Ver.--Ref. assesses refinement at fixed candidate counts; scaling varies $N$.

\section{Conclusion}
We introduce \textbf{SAGA}, an agentic framework that decouples frozen video
synthesis from perception-driven control, and \textbf{SurgControl-Bench} for
command-response, geometric, and appearance evaluation. The framework combines
modular verification, refinement, and cross-segment feedback to improve command
following and tool consistency without generator training. Independent TAPIR
and DINOv2 evaluation separates reported performance from the agent's selection
signals, characterizing verification trade-offs and refinement gains under
matched candidate budgets. These automated measurements do not establish
clinical utility or physical RCM compliance.

\clearpage

\section*{LLM Usage Statement}
We used OpenAI Codex to assist with experimental-code implementation
and debugging, and with drafting and language revision across the
manuscript. Codex also supported preliminary
discussions of experimental protocols and exploratory diagram drafts.
The authors determined the final experimental design and prepared the
final figures.

{
    \small
    \bibliographystyle{ieeenat_fullname}
    \bibliography{main}
}

\end{document}